\documentclass[lettersize,journal]{IEEEtran}

\usepackage[switch]{lineno} 
\usepackage{amsmath,amsfonts}
\usepackage{array}
\usepackage{textcomp}
\usepackage{url}
\usepackage{verbatim}
\usepackage{graphicx}
\usepackage{import}
\usepackage{color,soul}
\usepackage[table,dvipsnames]{xcolor}
\usepackage{subcaption}
\usepackage{tikz}
\newcommand*\circled[1]{\tikz[baseline=(char.base)]{
        \node[shape=circle,draw,inner sep=1.5pt, Black, fill=Black] (char)
               {\color{white}\scriptsize\textbf{#1}};}%
        }

\DeclareMathOperator*{\argmin}{arg\,min}

\usepackage{titlesec}
\title{Trajectory-to-Action Pipeline (TAP): Automated Scenario Description Extraction for Autonomous Vehicle Behavior Comparison}
\author{Aron Harder and Madhur Behl
\thanks{Department of Computer Science, University of Virginia, Charlottesville, Virginia, USA}
\thanks{
This work has been submitted to the IEEE for possible publication.
Copyright may be transferred without notice, after which this version may no longer be accessible.
}
}

\begin{document}

\maketitle

\begin{abstract}
Scenario Description Languages (SDLs) provide structured, interpretable embeddings that represent traffic scenarios encountered by autonomous vehicles (AVs), supporting key tasks such as scenario similarity searches and edge case detection for safety analysis.
This paper introduces the Trajectory-to-Action Pipeline (TAP), a scalable and automated method for extracting SDL labels from large trajectory datasets. 
TAP applies a rules-based cross-entropy optimization approach to learn parameters directly from data, enhancing generalization across diverse driving contexts. 
Using the Waymo Open Motion Dataset (WOMD), TAP achieves 30\% greater precision than Average Displacement Error (ADE) and 24\% over Dynamic Time Warping (DTW) in identifying behaviorally similar trajectories. Additionally, TAP enables automated detection of unique driving behaviors, streamlining safety evaluation processes for AV testing. 
This work provides a foundation for scalable scenario-based AV behavior analysis, with potential extensions for integrating multi-agent contexts.

\end{abstract}

\section{Introduction}
Autonomous vehicles (AVs) are nearing widespread deployment, but ensuring their safety remains a critical challenge. 
Public AV testing has revealed critical failures, sometimes leading to tragic outcomes~\cite{mccarthy2022autonomous, shepardson_2024, brown2023halting, bote_2023, song_2021, templeton_2023, o2016we, rodriguez_2023}. 
A methodical approach for measuring and improving safety for AVs is necessary to reduce these outcomes.
A central question in AV development and testing is: How do we know which autonomous vehicle (AV) is safer?
This fundamental question underlines much of the debate surrounding AV deployment and safety. 
Conventional metrics such as miles driven or disengagement counts provide a limited and often inadequate view of AV performance. 
Lower disengagements, for instance, does not necessarily equate to safer or more reliable driving behavior, as it fails to capture the full range of complex interactions AVs encounter in real-world traffic situations.
To objectively compare the safety of two AVs, we need a framework that goes beyond these basic metrics.

Comparing the safety of two AVs is challenging because they often operate under different conditions, different cities etc. 
Direct safety comparisons are unreliable without a common basis.
Rather, a more viable approach is to analyze AV behavior when exposed to similar traffic scenarios.
By evaluating AV performance in high-stakes situations, we gain deeper insight into their safety and robustness. However, implementing and scaling this approach requires the ability to automatically describe, interpret, and compare traffic scenarios across large datasets. 
This challenge motivates our work: How can we generate standardized, interpretable descriptions of traffic scenarios to identify both similar and rare situations for AV comparison?

Scenario Description Languages (SDLs) provide structured, interpretable descriptions of traffic scenarios by capturing both static (road geometry, traffic signals, etc.) and dynamic (vehicle maneuvers) elements.
SDLs can facilitate critical AV testing and safety tasks such as large-scale scenario discovery, real-to-sim transfers, post-deployment monitoring, and programmatic scenario generation.
Additionally, by encoding traffic scenarios into SDL embeddings, we can compare and identify similar scenarios allowing us to observe AVs in a similar traffic situation and, hence, reasoning about their relative safety.
Frameworks such as Scenic~\cite{Scenic} and OpenScenario~\cite{OpenScenario}, offer ways to define SDLs for these purposes. 

However, despite their utility, significant limitations of constructing SDLs remain: Due to the lack of ground-truth annotations, the extraction of semantic SDL labels from AV data is currently manual, requiring intensive oversight. 
Manual labeling of such data is impractical due to the vast scale of AV trajectory datasets. 
Since AV data sets do not provide labeled actions, using supervised learning techniques to extract SDLs is infeasible.
This bottleneck limits the scalability of SDLs for comprehensive scenario-based testing across diverse AV datasets.
While advances in computer vision have made static element extraction (such as road geometry) more feasible~\cite{segment_anything,SemSeg_Transformer}, extracting dynamic elements - particularly vehicle behaviors - is far more complex.
Accurately capturing spatio-temporal actions requires more than video-based similarity; as it involves understanding nuanced trajectory patterns. 

To overcome these limitations, we introduce the Trajectory-to-Action Pipeline (TAP), a method designed to automate the extraction of interpretable vehicle behavior SDL descriptions from unlabeled AV trajectory data. 
TAP uses cross-entropy optimization to automatically derive SDL labels based on the dataset’s distribution. 
This automated approach addresses the pressing need for scalable, interpretable scenario descriptions, enabling TAP to compare vehicle behaviors and identify unique vehicle behaviors without the need for manual input. The contributions of this paper are as follows:
\begin{enumerate}
    \item We present the Trajectory-to-Action Pipeline (TAP), an automated method for extracting interpretable vehicle behavior SDL embeddings from trajectory data, reducing reliance on manual labeling and enabling scalable scenario analysis.
    \item We validate TAP on the Waymo Open Motion Dataset, demonstrating its precision in identifying behaviorally similar trajectories and automatic detection of unique scenarios.
\end{enumerate}

By automating SDL extraction, TAP provides a robust, and adaptable foundation for large-scale AV scenario and trajectory data analysis, that could ultimately help support effective AV safety evaluations in real-world environments.

\section{Related Work}
Our research intersects with scenario description languages, scene understanding, and trajectory similarity. 

\noindent \textbf{Scenario Description Languages}
Scenario Description Languages (SDLs) serve as high-level embeddings to describe traffic scenarios, capturing key elements like vehicles, pedestrians, and traffic lights. Existing SDLs such as Fortellix’s MSDL~\cite{M-SDL}, Scenic~\cite{Scenic}, and OpenScenario~\cite{OpenScenario} focus on comprehensive scene descriptions of environment, road geometry etc.
However, our work diverges by concentrating specifically on describing autonomous vehicle motion to facilitate behavior comparison.
Previous methods, such as Scenario2Vector~\cite{aron_date24, 
aron_iccps21}, have explored using SDLs for scenario similarity using video and natural language data. 
In contrast, our approach differs by introducing an automated pipeline for extracting vehicle motion - focused SDL labels, enhancing scalability and precision.
Alternative approaches include graph-based representations like 
Roadscene2vec and GraphAD~\cite{Roadscene2vec, GraphAD, PreGSU}, 
which model scenes using nodes for agents and edges for their 
interactions. 
While effective for capturing vehicle interactions, these methods struggle to accurately represent vehicle motion dynamics.
Recently, natural language descriptions~\cite{scene_understanding_captions, 
explainable_cars} have been used to capture complex scene interactions. 
Although expressive, this approach requires extensive training data and includes extraneous linguistic elements, making it less efficient for scenario comparison. 
Another approach, used by works such as iMotion~\cite{iMotion} and MotionLM~\cite{motionLM}, uses Large Language Models to describe scenes. However, iMotion creates their prompts using template sentences, making it just another SDL. MotionLM treats motion tokens as a language, which loses the interpretability of a traditional SDL.
In contrast, our method achieves a high information density while maintaining interpretability, focusing on the essential motion characteristics relevant to vehicle behavior.


\noindent \textbf{Scene Understanding}
Scene understanding is a vast field within computer vision, of which traffic scene understanding is a small portion.
In \cite{nlp_video_retrieval1}, researchers use dashcam video for scenario retrieval.
Other scene understanding work~\cite{SpatialCNN_SceneUnderstanding,RoadSceneUnderstanding,Vid2Bev} focuses on modeling the relationship between road agents and their environment features like lanes, etc.  
For instance, Vid2Bev~\cite{Vid2Bev} converts camera footage into a bird’s-eye view, enabling AVs to create a vectorized world for planning and control.
These algorithms are important for AVs to create a vectorized view of the world in which their planning and control modules operate.
Since AV planning and control modules already operate in this vectorized space, we leverage the existing data available to the vehicle. This allows us to analyze vehicle behavior as perceived by the vehicle itself, rather than relying on camera-based perception. Thus, our focus is on analyzing vectorized vehicle data, bypassing the need to re-implement perception algorithms.


\noindent \textbf{Trajectory Similarity}
In the vectorized world used by AVs, vehicle behavior is represented by trajectories of positional history. 
Traditional methods for trajectory similarity, such as Average Displacement Error (ADE)~\cite{ADE} and Dynamic Time Warping (DTW)~\cite{DTW}, compare distances between corresponding points. 
However, these methods struggle with varying trajectory lengths, requiring resampling, which can strip away crucial information. 
Deep learning methods~\cite{dnn_traj_sim} are more robust to length differences but remain sensitive to early deviations, leading to accumulated errors. 
In contrast, our Trajectory-to-Action Pipeline (TAP) uses Scenario Description Language (SDL) labels to capture interpretable action-level behaviors, preserving essential motion characteristics and ensuring robustness to trajectory variations.

\section{Scenario Description Language}
\label{sec:sdl}

To evaluate and compare autonomous vehicle behaviors effectively, it is essential to represent driving scenarios in a structured format that captures the complexities of vehicle actions over time. 
Scenario Description Languages (SDLs) fulfill this need by encoding static elements such as intersections and dynamic elements such as vehicles and pedestrians.
These elements are easily obtained from the vectorized data used by AVs, as the output of the perception module.
Therefore, we focus on the temporal task of describing trajectories as sequences of interpretable actions.
This represents a more difficult task, as vectorized data does not contain labels for AV behavior.
Therefore, we introduce SDL labels that represent a vehicle's behavior in terms of discrete, high-level actions, making it possible to analyze scenarios and compare trajectories based on their behavioral patterns. 

\subsection{Overview of SDL Structure}

\begin{figure}
    \centering
    \includegraphics[width=\columnwidth]{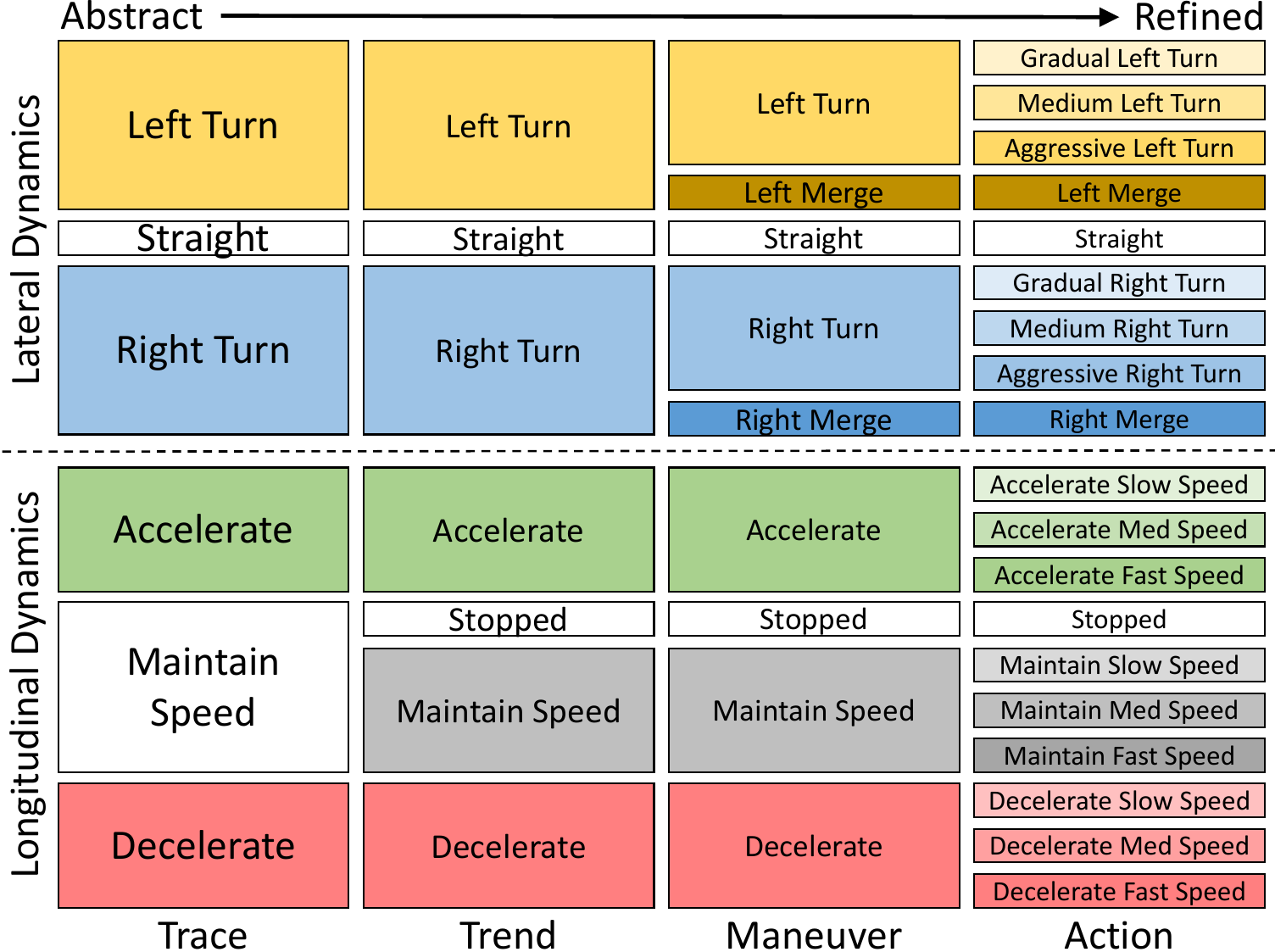}
    \caption{Actions in our SDL, organized along a spectrum of complexity. At the abstract level, there are three lateral and three longitudinal labels, which can be further subdivided into finer-grained labels for more detailed analysis.}
    \label{fig:action_spectrum}
\end{figure}

SDLs provide a systematic way to describe driving behaviors.
Based on the SDL framework outlined by~\cite{aron_date24}, we define an SDL that characterizes autonomous vehicle behaviors through a defined set of temporal actions.
Figure~\ref{fig:action_spectrum} illustrates the set of vehicle behaviors employed in our SDL framework.
We partition the driving behavior of the autonomous vehicle in terms of lateral and longitudinal actions, where lateral actions capture directional movements, such as turns or lane changes and longitudinal actions capture changes in speed, such as acceleration or deceleration.
The separation of lateral and longitudinal actions in SDL reflects the control systems in autonomous vehicles, allowing flexible and accurate representation of combined driving behaviors. 
This structure also aligns with established practices in vehicle dynamics, and allows us to evaluate autonomous vehicle responses across scenarios in a way that is 
interpretable, consistent, and comparable.

\subsection{Hierarchical Levels of SDL Labels}
\label{sec:sdl_hierarchy}

As shown in Figure~\ref{fig:action_spectrum} the SDL uses a hierarchical approach, with multiple levels of refinement that capture vehicle behavior at varying 
degrees of detail. 
These levels, progressing from the simplest to the most detailed, enable SDL to adapt to different levels 
of data availability and task requirements. 
Below, we define each level in the hierarchy and describe how it represents driving behavior.

\paragraph{Trace Level}
At the most abstract level, the Trace level encodes actions using broad categories. Lateral actions are simplified to 
basic movement types (e.g., "Left Turn," "Right Turn," "Straight"), while longitudinal actions are captured as "Accelerate," 
"Decelerate," or "Maintain Speed." This level provides a coarse-grained view of driving behavior, suitable for high-level analysis.

\paragraph{Trend Level}
The Trend level refines the Trace level by distinguishing between different speed and state categories. For example, 
the "Maintain Speed" action is divided into "Stopped" and "Moving" states. This level provides a more nuanced understanding 
of vehicle behavior, capturing the persistence of actions over time.

\paragraph{Maneuver Level}
At the Maneuver level, SDL labels begin to capture more complex actions, particularly in terms of lateral behavior. 
For example, lane-change maneuvers such as "Left Merge" and "Right Merge" are explicitly included. This level is valuable 
for identifying interactions between vehicles or with the road environment, as it captures intentional, goal-directed behaviors.

\paragraph{Action Level}
The most detailed level, the Action level, distinguishes actions based on the intensity and type of movement. 
For instance, turns are classified into "Gradual," "Medium," and "Aggressive," depending on the vehicle’s yaw rate. 
Similarly, accelerations and decelerations are categorized by speed (e.g., "Accelerate Slow," "Accelerate Medium"). 
The Action level is particularly useful for granular analyses, where precise distinctions in vehicle dynamics are required.

The hierarchy of SDL levels allows flexibility in matching the level of detail to the data and task requirements. 
For example, a high-level comparison across many scenarios may only require Trace-level labels, while a detailed analysis of specific behaviors benefits from Action-level granularity.
Each of these levels provides a set of possible labels for lateral and longitudinal components, which we denote as 
\(L_{\text{lat}}\) and \(L_{\text{long}}\), respectively. At any level \(X\), the sets \(L_{\text{lat}}^{X}\) and 
\(L_{\text{long}}^{X}\) define the allowable actions for lateral and longitudinal behaviors. For example:
\[
L_{\text{lat}}^{\text{Trend}} = \{\text{Left Turn}, \text{Right Turn}, \text{Straight}\}
\]
\[
L_{\text{long}}^{\text{Trend}} = \{\text{Accelerate},\text{Stopped}, \text{Maintain Speed}, \text{Decelerate}\}
\]
The transformation from a trajectory to SDL labels involves segmenting the trajectory into intervals and assigning actions based on the vehicle’s lateral and longitudinal states. Thresholds defined for yaw rate, acceleration, and velocity determine the classification at each level. 
In our work, SDL labels describe behaviors that are at least 1 second in length to ensure that the vehicle is exhibiting an intentional behavior, rather than noise.

\section{Problem Statement}
\label{sec:problem_statement}

Our objective is to automatically extract SDL embeddings from scenario data.
Let $S_i \in S$ be a scenario out of dataset of scenarios $S$.
Each scenario $S_i$ contains a set of $k$ vehicles $\{V^i_1,V^i_2,\dots,V^i_j,\dots,V^i_k\}$.
The motion of vehicle $V^i_j$ is captured by a set of vehicle states $s = \{x,y,v,a,\phi,\omega\}$ where $x$ and $y$ are the vehicle's position in the map frame, $v$ is the vehicle's velocity in m/s along the heading $\phi$ (radians), $a$ is the vehicle's acceleration in m/s$^2$, and $\omega$ is the vehicle's yaw rate in rad/s.
For vehicle $V^i_j$, trajectory $\tau^j = \{s^j_1,s^j_2,\dots,s^j_T\}$ is an ordered sequence of the vehicle's states across the duration $T$ of the scenario.
To define the scope and limitations of our framework, we make the following assumptions.
\noindent \textbf{List of Assumptions:}
\begin{enumerate}
    \item Although scenarios typically include static elements (e.g., road geometry), our framework focuses on extracting vehicle behavior from trajectory data. We assume static elements can be extracted separately using computer vision.
    \item While real-world traffic involves agent interactions, this work isolates ego trajectories to demonstrate the feasibility of encoding vehicle behavior using SDLs. Future work can extend this to multi-agent scenarios~\cite{causal_agents}.
    \item We use the hierarchical SDL schema described in Section~\ref{sec:sdl}, though the method can adapt to frameworks like Scenic and OpenScenario.
    \item We assume no ground truth action labels are provided for the trajectory. Without labeled data, supervised approaches like neural networks are infeasible. Unsupervised methods (e.g., k-means) rely on measures such as ADE, which inadequately capture spatio-temporal behavior (Section~\ref{sec:result2}).
\end{enumerate}

Under these assumptions, given a trajectory $\tau^j$, our goal is to find a function
\begin{equation}
    f_{\Theta}: \tau^j \rightarrow L^i_j
    \label{eqn:sdl_extraction}
\end{equation}
that defines a mapping between an input vehicle trajectory $\tau^j$, and produces an SDL label $L^i_j$ as the output describing the vehicle's behavior.
The SDL label $L^i_j = \langle \{L_{\text{lat},1},L_{\text{lat},2},\dots\},\{L_{\text{long},1},L_{\text{long},2},\dots\} \rangle$ is a tuple containing a chronologically ordered sequence of the lateral and longitudinal SDL labels as defined in Section~\ref{sec:sdl_hierarchy}.
$f_{\Theta}$ is realized through a set of rules, where $\Theta$ denotes a set of separation thresholds.
The problem of automatic SDL extraction is the problem of determining values for these thresholds.

\subsection{Threshold Calculation Problem}
\label{sec:problem_threshold}

The objective of threshold calculation is to find separation thresholds $\Theta$ that ensure samples with the same label are more similar to each other than to those with different labels. Automatically deriving these thresholds from the data enables our method to generalize across datasets. 
Let $D$ denote the set $D = \{D_{\omega}, D_a, D_v\}$ of distributions of 1 second sampled averages of yaw rate $\omega$, acceleration $a$, and velocity $v$ across the entire dataset, spanning all scenarios.
For example, vehicle $V^i_j$ contributes the 1 second sampled average of its accelerations $\{\bar{a}^i_{j,1},\bar{a}^i_{j,2},\dots\}$ to the distribution $D_a$.
The sampling interval of 1 second allows noise to be smoothed without losing key behavioral differences.
The velocity distribution $D_v$ and yaw rate distribution $D_{\omega}$ are defined similarly.
The separating thresholds $\Theta = \{\Theta_{\omega},\Theta_a,\Theta_v\}$ are defined over their corresponding distributions $D_{\omega}$, $D_a$, and $D_v$, where $\Theta_{\omega} = \{\theta_{\omega,\text{str}},\theta_{\omega,\text{grad}},\theta_{\omega,\text{med}}\}$, $\Theta_{a} = \{\theta_{a,\text{dec}},\theta_{a,\text{acc}}\}$, and $\Theta_{v} = \{\theta_{v,\text{stop}}, \theta_{v,\text{slow}},\theta_{v,\text{med}}\}$ are thresholds that realize the rule-based function $f_{\Theta}$.
For each distribution, the corresponding thresholds separate the distribution into distinct partitions that map vehicle states to labels.
E.g. acceleration thresholds $\Theta_a$ separate the acceleration distribution $D_a$ into partitions: Decelerate, Maintain Speed, and Accelerate.
The full list of partitions and the logical rules are included in Table~\ref{table:partitions}.
Each sample that falls into the same partition is given the same SDL label.
The distance between two samples within a partition $p1,p2 \in D_{\text{part}}$ is the absolute value of their difference $|p1-p2|$ as a measure of their similarity.

\begin{table}[b]
\centering
\renewcommand{\arraystretch}{1}
\begin{tabular}{ |c|c| }
 \hline
 \rowcolor[gray]{0.9} \textbf{Logical Rule} & \textbf{Partition} \\
 \hline\hline
 \multicolumn{2}{|c|}{\textbf{Yaw Rate ($\omega$)}} \\
 \hline
 $|\omega| \leq \theta_{\omega,\text{str}}$ & Straight \\
 $\theta_{\omega,\text{str}} < |\omega| \leq \theta_{\omega,\text{grad}}$ & Gradual Turn \\
 $\theta_{\omega,\text{grad}} < |\omega| \leq \theta_{\omega,\text{med}}$ & Medium Turn \\
 $\theta_{\omega,\text{med}} < |\omega|$ & Aggressive Turn \\
 \hline
 \multicolumn{2}{|c|}{\textbf{Acceleration ($a$)}} \\
 \hline
 $a \leq \theta_{a,\text{dec}}$ & Decelerate \\
 $\theta_{a,\text{dec}} < a \leq \theta_{a,\text{acc}}$ & Maintain Speed \\
 $\theta_{a,\text{acc}} < a$ & Accelerate \\
 \hline
 \multicolumn{2}{|c|}{\textbf{Velocity ($v$)}} \\
 \hline
 $v \leq \theta_{v,\text{stop}}$ & Stopped \\
 $\theta_{v,\text{stop}} < v \leq \theta_{v,\text{slow}}$ & Slow \\
 $\theta_{v,\text{slow}} < v \leq \theta_{v,\text{med}}$ & Medium \\
 $\theta_{v,\text{med}} < v$ & Fast \\
 \hline
\end{tabular}
\caption{
The rule-based criteria for data partitions. For instance, vehicle acceleration is separated by $\theta_{a,\text{acc}}$ and $\theta_{a,\text{dec}}$ into partitions of Decelerate, Maintain Speed, and Accelerate. 
}
\label{table:partitions}
\end{table}

For each partition, $D_{\text{part}}$, the intra-partition similarity is measured using the average pairwise distance: 
\begin{align}
  \mu_{\text{part}} = \frac{\sum_{p1,p2 \in D_{\text{part}}}|p1-p2|}{{|D_{\text{part}}| \choose 2}}.   
\end{align}
This value is calculated as the sum of all pairwise distances between samples within the partition, divided by the total number of sample pairs. 
A lower value for $\mu_{\text{part}}$ indicates higher similarity among the samples in that partition.
Given two partitions $\text{part1},\text{part2} \in D$ we can now define the inter-partition similarity as $\mu_{\text{part1}}-\mu_{\text{part2}}$.
Therefore, objective function
\begin{align}
   J(\Theta) = \sum_{part1,part2 \in D} ||\mu_{\text{part1}}-\mu_{\text{part2}}||^2 
\end{align}
computes the squared inter-partition distances that measure the similarity between different partitions.
Minimizing the value of $J(\Theta)$ means that each of the partitions will be equally similar.
Our goal then is to find thresholds that minimize the value of $J(\Theta)$
\begin{equation}
    \Theta^{*} = \argmin_{\Theta} J(\Theta)
    \label{eqn:cem_objective}
\end{equation}

The benefit of analyzing lateral and longitudinal distributions separately, is that the thresholds $\Theta_{\omega}$, $\Theta_a$, and $\Theta_v$ can be minimized independently.
For instance, $\Theta^{*}_{a} = \argmin_{\Theta_{a}} J(\Theta_{a})$.

\subsection{SDL Similarity Search Problem}
\label{sec:problem_sim}

Given a set of labeled trajectories $S^L$ and a reference label $L_{ref} = \langle \{L_{\text{lat},1},L_{\text{lat},2},\dots\},\{L_{\text{long},1},L_{\text{long},2},\dots\} \rangle$, the objective of the SDL similarity search problem is to find the set of labeled trajectories $S^L_{\text{ref}} \subset S^L$ that are similar to the reference label.



The similarity between SDL labels is given by the distance function $g(L_{\text{ref}},L^i_j) = d$ where $d \in \mathbb{R}$ is a scalar distance and $L^i_j$ is the SDL label that describes the behavior of vehicle $V^i_j$.
Therefore, the similarity search problem is to construct the set
\begin{equation}
    S^L_{\text{ref}} = \{L^i_j: g(L_{\text{ref}},L^i_j) < d_{\text{sim}}\}
    \label{eqn:sim_search}
\end{equation}
where all labels $L^i_j$ are within a distance $d_{\text{sim}}$ of the reference label.
Setting $d_{\text{sim}} = 0$ requires that all similar samples $L^i_j \in S^L_{\text{ref}}$ have the exact same lateral and longitudinal label as the reference.

\textbf{Identifying Unique Scenarios}
The similarity search can also automatically identify unique scenarios in a dataset.
Here, a unique scenario is a scenario that is not encountered anywhere else in the dataset.
Unique behavior can be identified from unique SDL labels. 
It follows from Equation~\ref{eqn:sim_search} that if the set of similar labels $S^L_{\text{ref}} = \{\emptyset\}$ is empty, then the reference label itself is unique, indicating that the vehicle's behavior is found nowhere else in the dataset.

\section{Methodology}
\begin{figure}
    \centering
    \includegraphics[width=0.9\linewidth]{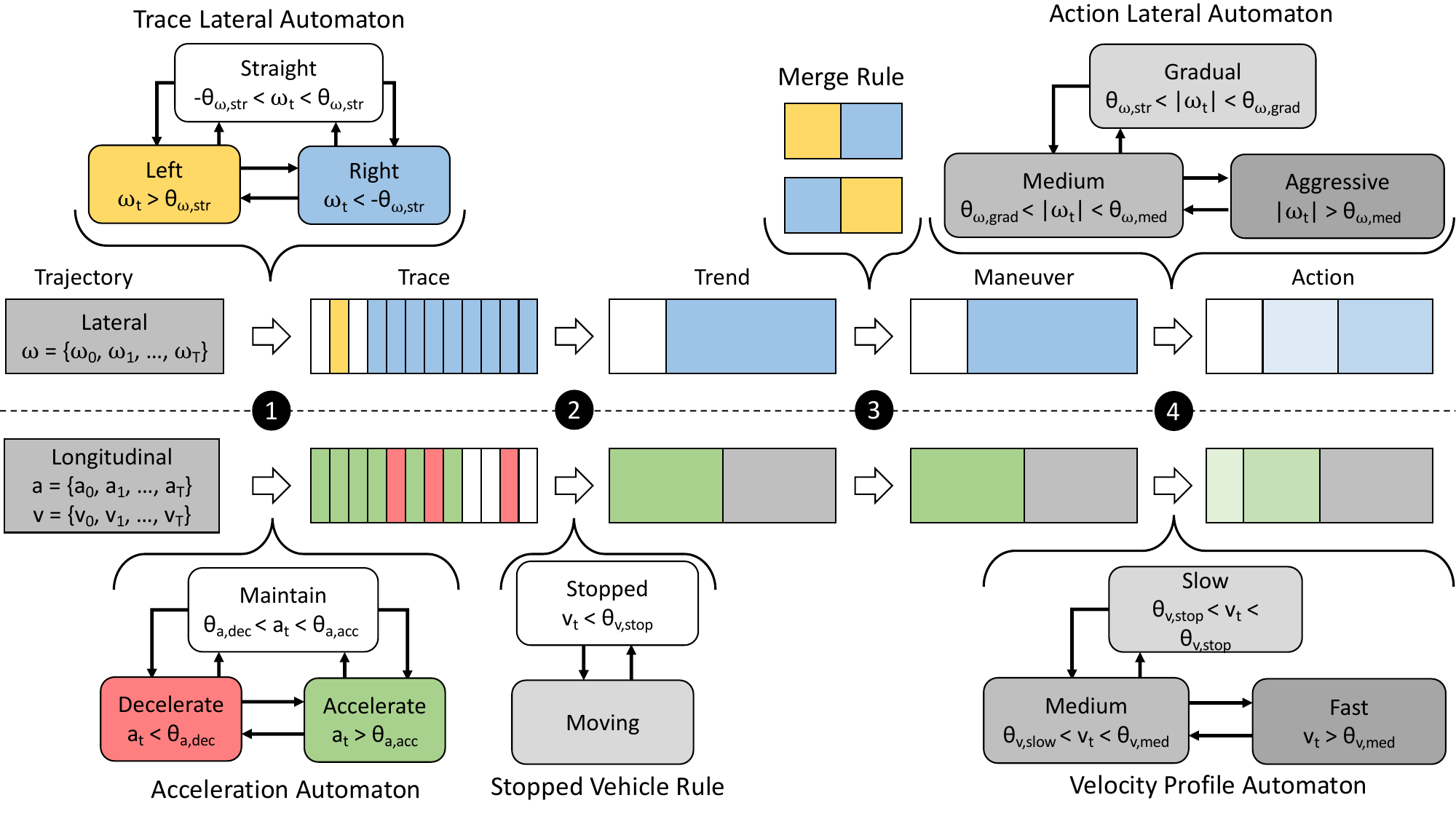}
    \caption{Trajectory-to-Action Pipeline (TAP) starts with acceleration, velocity, and yaw rate data as input. At each stage, it uses automata and rules to produce SDL labels at the level of refinement corresponding to that stage. The final stage of TAP produces action-level SDL labels for every vehicle.}
    \label{fig:sdl_pipeline}
\end{figure}
The goal of the Trajectory-to-Action Pipeline (TAP) is to automate the extraction of Scenario Description Language (SDL) labels from raw, unlabeled vehicle trajectory data.
The absence of ground truth labels prevents the direct use of supervised learning approaches. 
TAP extracts SDL labels offline, for post hoc analysis of vehicle behavior.
Figure~\ref{fig:sdl_pipeline} provides an overview of the entire Trajectory-to-Action Pipeline.
The pipeline consists of multiple stages, each refining the classification of vehicle actions. 
The implementation of TAP will be released on GitHub upon the paper’s publication.
Below, we detail the different stages involved in TAP.

\subsection{Trajectory-to-Action Pipeline}
\label{sec:methodology_tap}

Given a trajectory $\tau^j$ as input, TAP uses the yaw rate $\omega$ as lateral input and acceleration $a$ and velocity $v$ as longitudinal inputs.
Our SDL defines separate lateral and longitudinal labels, and TAP enforces this separation in its analysis of the data.
Starting from the input trajectory data, TAP involves a series of rules in 4 stages to go from the most abstract, trace-level SDL labels in stage 1 to the most refined action-level SDL labels in stage 4.
Figure~\ref{fig:sdl_pipeline} shows the different stages of TAP which are described in detail below.


\textbf{Stage 1 - Trace}
The input to the first stage of TAP is a raw trajectory with yaw rate $\omega$ and acceleration $a$.
The goal of stage 1 is to produce trace-level SDL labels $\{L^{\text{Trace}}_{\text{lat},1},L^{\text{Trace}}_{\text{lat},2},\dots,L^{\text{Trace}}_{\text{lat},T}\}$ and $\{L^{\text{Trace}}_{\text{long},1},L^{\text{Trace}}_{\text{long},2},\dots,L^{\text{Trace}}_{\text{long},T}\}$ to describe vehicle behavior, where $L^{\text{Trace}}_{\text{lat}}$ are \{Left Turn, Right Turn, Straight\}, while $L^{\text{Trace}}_{\text{long}}$ are \{Accelerate, Decelerate, Maintain Speed\}.
TAP utilizes lateral and longitudinal rules-based automata to automatically extract trace-level SDL labels.
The trace lateral automaton takes as input the trajectory's yaw rate $\omega = \{\omega_0,\omega_1,\dots,\omega_T\}$.
The automaton has three nodes, Left, Right, and Straight, that correspond to the three trace-level lateral SDL labels.
The rules-based criteria for the trace lateral automaton (shown in Figure~\ref{fig:sdl_pipeline}) is given by:
\begin{equation}
    \begin{cases}
        \omega_t > \theta_{\omega,\text{str}}\,; & L^{\text{Trace}}_{\text{lat}} = \text{Left} \\
        -\theta_{\omega,\text{str}} < \omega_t < \theta_{\omega,\text{str}}\,; & L^{\text{Trace}}_{\text{lat}} = \text{Straight} \\
        \omega_t < -\theta_{\omega,\text{str}}\,; & L^{\text{Trace}}_{\text{lat}} = \text{Right}
    \end{cases}
\end{equation}
Starting at $t=0$, each yaw rate measurement $\omega_t$ from the trajectory is fed into the trace lateral automaton to produce a label $L^{\text{Trace}}_{\text{lat},t}$

\begin{figure}
    \centering
    \includegraphics[width=1.0\linewidth]{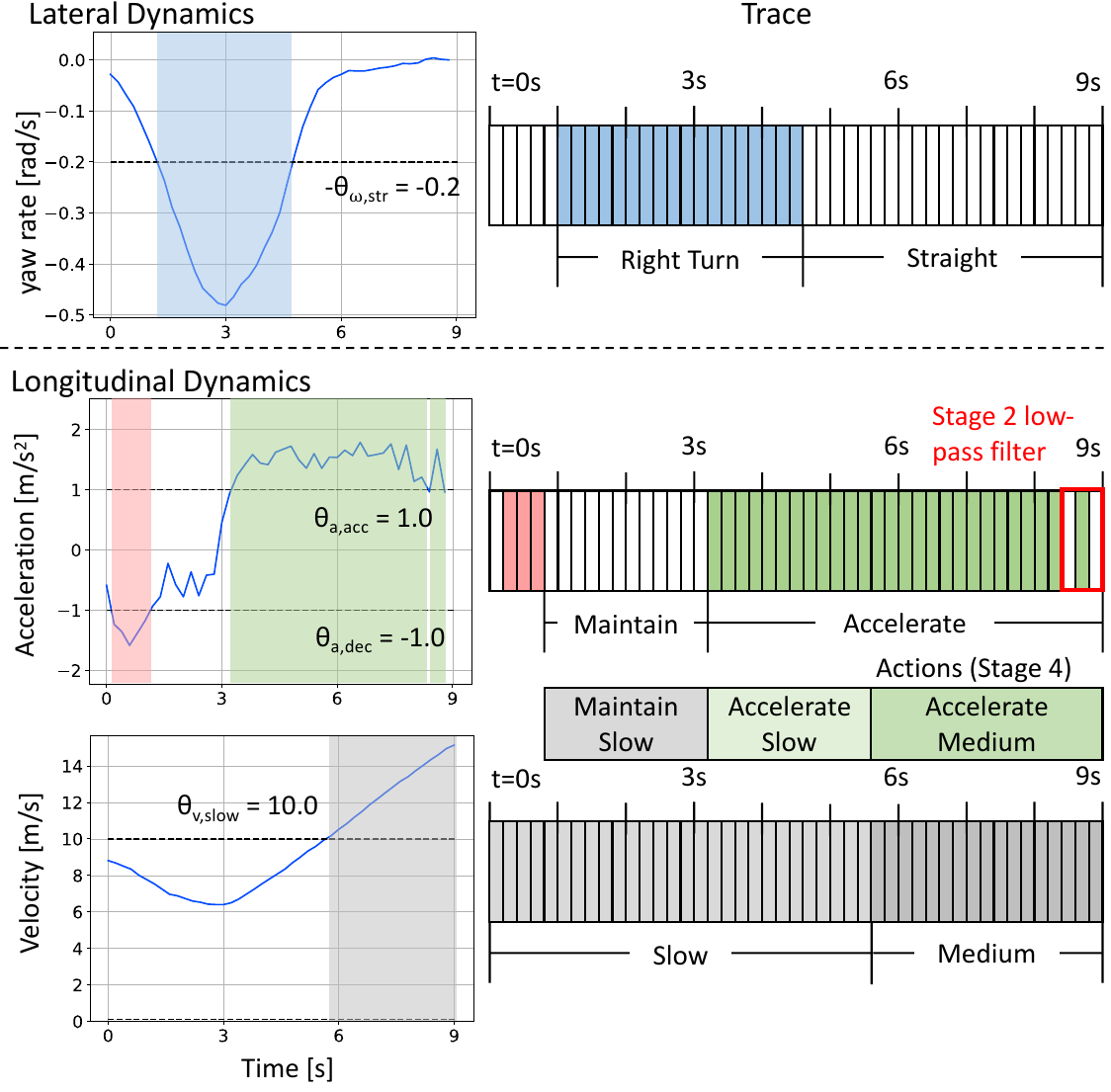}
    \caption{Example trajectory of a vehicle making a right turn, with yaw rate $\omega$, acceleration $a$, and velocity $v$ analyzed by TAP. Lateral SDL labels (trace-level) are Right Turn and Straight, while longitudinal labels are Maintaining Speed and Accelerating. At stage 4, TAP refines the velocity data into SDL actions: Maintain Slow Speed, Accelerate Slow Speed, and Accelerate Medium Speed.}
    \label{fig:example_traj}
\end{figure}

Consider the example shown in Figure~\ref{fig:example_traj}.
This figure shows the yaw rate $\omega$, acceleration $a$, and velocity $v$ at each time step for an example trajectory.
At $t = 0.0$ seconds, the yaw rate satisfies the criteria for the trace-level SDL label Straight.
At $t = 1.0$ seconds, the yaw rate falls below the threshold $-\theta_{\omega,\text{str}}$.
Therefore, the extracted label is Right Turn as the automaton transitions to the Right node.
At $t = 4.6$ seconds, the yaw rate rises back above the threshold to again be assigned a Straight SDL label.
Similar to the trace lateral automaton, the acceleration automaton produces trace-level longitudinal SDL labels $L^{\text{Trace}}_{\text{long}}$.
Figure~\ref{fig:example_traj} also shows how the acceleration automaton produces longitudinal labels from the acceleration $a$ for our example.
At the end of stage 1, TAP has automatically extracted trace-level SDL labels $L^{\text{Trace}}$ from trajectory yaw rate $\omega$ and acceleration $a$.

\textbf{Stage 2 - Trend}
The only difference between trace-level and trend-level SDL labels is in the longitudinal aspect, where a Stop label is introduced into the SDL.
The input to stage 2 of TAP is a trajectory with trace-level SDL labels $L^{\text{Trace}}$ as well as velocity $v = \{v_0,v_1,\dots,v_T\}$ which has not been used thus far.
The goal of stage 2 is to detect when the vehicle is stopped as well as handle noise in the data.
Whether or not a vehicle is stopped cannot be determined by the acceleration automaton.
Therefore, at this stage, TAP implements a Stopped Vehicle Rule, governed by threshold $\theta_{v,\text{stop}}$, to analyze the velocity $v_t$ and detect instances when the vehicle is not moving.
When $v_t < \theta_{v,\text{stop}}$, TAP assigns an action of Stopped to the label $L^{\text{Trend}}_{\text{long},t}$.
It is possible that when the vehicle is stopped, the data contains non-zero yaw rate $\omega$ values. We postulate that this is due to noise, as a vehicle that is not moving cannot be turning.
For this reason, at any time $t$ where the vehicle is stopped, the trend-level lateral SDL label $L^{\text{Trend}}_{\text{lat},t}$ is assigned the Straight label.
During the trend stage, TAP also utilizes a low-pass filter to smooth the data.
If there are only a few frames with a label that is not consistent with the neighboring frames, TAP computes the average value to smooth these outliers.
Going back to our running example in Figure~\ref{fig:example_traj}, the last few frames of the acceleration's trace-level SDL label are very short in duration but got assigned a Maintain Speed label, despite being surrounded by Accelerate labels.
These would get smoothed by the low-pass filter to all have the trend-level SDL label of Accelerate.
At the end of the trend stage, TAP has added the Stop SDL label and removed noise to produce trend-level SDL labels $L^{\text{Trend}}$.



\textbf{Stage 3 - Maneuver}
At the maneuver-level in the SDL hierarchy, the only change is the addition of a new lateral behavior called the Merge maneuver.
A Merge is a complex behavior and is defined by a turn in one direction followed by a turn in the opposite direction.
Vehicles typically experience Merges when changing lanes or entering/exiting a freeway.
Merges are difficult to detect because the trend-level lateral SDL labels only partition lateral behavior into a Left Turn, Right Turn, or Straight, whereas a Merge can be composed of any combination of these three SDL labels.
To detect Merges from trend-level lateral labels, TAP analyzes $L^{\text{Trend}}_{\text{lat}}$ to search for trajectories which feature turns in one direction, followed by a turn in the opposite direction.
In order to distinguish a Merge maneuver from successive Turn maneuvers, TAP enforces a time limit on the Merge.
This time limit dictates the maximum time interval between two turns in opposite directions.
Once the initial turn has ended, the turn back in the opposite direction must start within this time limit to be considered a Merge.
Since the occurrence of Merges was rare in our dataset (0.8\%), we manually examined the merge trajectories and chose 4 seconds as the maximum interval between turns in opposite directions.
The output of stage 3 of TAP is the maneuver-level SDL labels $L^{\text{Maneuver}}$

\textbf{Stage 4 - Action}
At the beginning of stage 4, every trajectory $\tau^j$ has maneuver-level SDL labels $L^{\text{Maneuver}}_{\text{lat}} = \{\text{Left Turn},\allowbreak \text{Left Merge}, \allowbreak \text{Straight}, \allowbreak \text{Right Turn},\allowbreak \text{Right Merge}\}$ and $L^{\text{Maneuver}}_{\text{long}} = \{\text{Accelerate}, \allowbreak \text{Stopped},\allowbreak \text{Maintain Speed},\allowbreak \text{Decelerate}\}$.
In the Action stage, the goal is to distill the maneuver-level SDL labels into more refined action-level SDL labels $L^{\text{Action}}$ that can capture even more nuanced lateral and longitudinal behaviors.
Not all turns are equal—some are gradual, while others are sharp. 
To capture this, action-level SDL labels classify turns as Gradual Turn, Medium Turn, and Aggressive Turn. 
Similarly, longitudinal actions like Accelerate, Maintain Speed, and Decelerate are further divided into speed profiles: Slow, Medium, and Fast.
To refine lateral maneuvers, TAP uses the action lateral automaton with thresholds \(\theta_{\omega,\text{str}}\), \(\theta_{\omega,\text{grad}}\), and \(\theta_{\omega,\text{med}}\) as defined in Table~\ref{table:partitions}. 
This automaton refines the maneuver-level labels into action-level SDL labels such as Gradual, Medium, and Aggressive Turns. 
Given a turn maneuver \(L^{\text{Maneuver}}_{\text{lat},t}\) and yaw rate \(\omega_t\), the output is the action-level label \(L^{\text{Action}}_{\text{lat},t}\).
During this stage, TAP also uses the velocity profile automaton which has nodes of Slow, Medium, and Fast.
The input to the velocity profile automaton is a longitudinal maneuver $L^{\text{Maneuver}}_{\text{long},t}$ and velocity $v_t$, and it assigns one of three different velocity profile labels to produce the action-level longitudinal SDL label $L^{\text{Action}}_{\text{long},t}$.
Referring back to Figure~\ref{fig:example_traj}, the bottom plot shows the vehicle's velocity. 
It starts at a Slow speed (14 mph) at \( t = 0.0 \) seconds and reaches Medium speed (22 mph) by \( t = 5.6 \) seconds. In the previous TAP stage, two longitudinal maneuvers - Maintain Speed and Accelerate - were identified. 
The velocity profile automaton further refines these into Maintain Slow Speed, Accelerate Slow Speed, and Accelerate Medium Speed.
The original maneuvers can sometimes be split into action-level labels shorter than 1 second. 
This poses a risk of data loss, as our SDL requires actions to be at least 1 second long. 
To address this, TAP assigns a single action based on the average value of the maneuver when the resulting actions are too short.
After the action stage, every trajectory $\tau^j$ has been labeled with the lateral actions $L^{\text{Action}}_{\text{lat}}$ and the longitudinal actions $L^{\text{Action}}_{\text{long}}$ which are the most refined SDL labels.

In summary, for each of the stages in TAP, the output corresponds to the ordered sequence of labels $L^{\text{X}}_{\text{lat}}$ and $L^{\text{X}}_{\text{long}}$ at increasing levels of refinement.
These labels are all the information necessary to construct the SDL label desribed in Section~\ref{sec:sdl}: $L^i_j = \langle \{L_{\text{lat},1},L_{\text{lat},2},\dots\}$,$\{L_{\text{long},1},L_{\text{long},2},\dots\} \rangle$ that capture the behavior of vehicle $V^{i}_{j}$ in scenario $S_i$.

\subsection{Optimizing SDL Extraction Thresholds with Cross-Entropy}
\label{sec:ce_methods}

The accuracy of the SDL labels generated by TAP depends on the separation thresholds $\Theta$ which are used in the transition rules of the different TAP automata.
The objective then is to automatically determine thresholds $\Theta$ from the distributions $D_{\omega}$, $D_a$, and $D_v$. 
This was described in Section~\ref{sec:problem_sim} as the minimization optimization problem for $J(\Theta)$ (Equation~\ref{eqn:cem_objective}).
The optimization problem is a search over the manifold of $J(\Theta)$ to find thresholds $\Theta^*$ that minimize the objective function.
For e.g. $J(\Theta_{\omega})$ is a 3-dimensional manifold over the yaw rate thresholds $\Theta_{\omega} = \{\theta_{\omega,\text{str}},\theta_{\omega,\text{grad}},\theta_{\omega,\text{med}}\} \in \mathbb{R}^3$.
If $J(\Theta)$ were a smooth function, this problem is trivially solved by computing the gradient $\nabla J(\Theta)$.
However, in our case, $J(\Theta)$ is defined by a set of disjointed rules (Table~\ref{table:partitions}), which in-turn depend on the thresholds.
As a result, due to $J(\Theta)$ non-smooth nature it is not differentiable, making $\nabla J(\Theta)$ unavailable in closed-form.
\begin{figure*}
    \centering
    \includegraphics[width=0.95\linewidth]{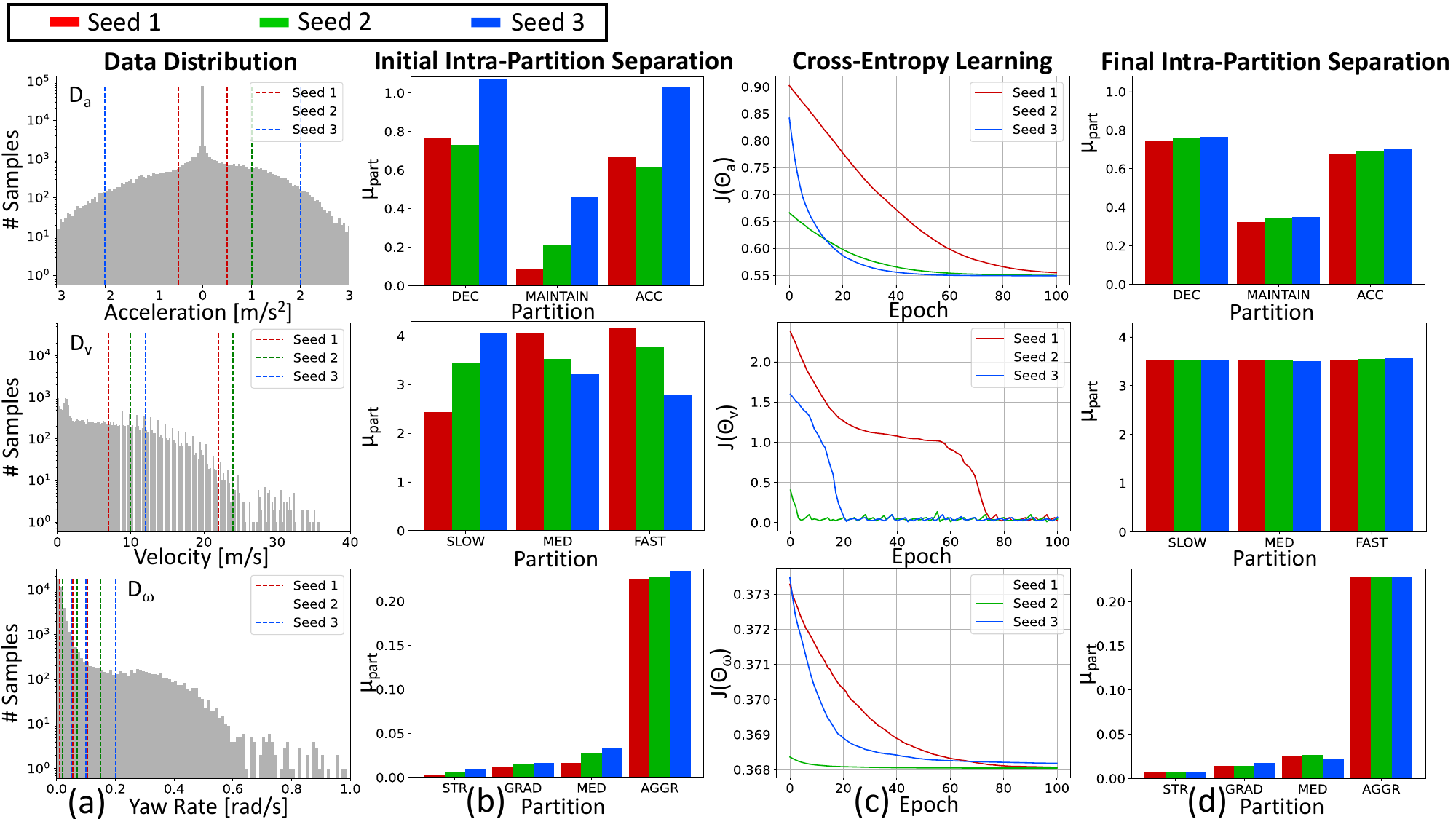}
   \caption{(a) Data distributions with initial seeds shown as dotted lines. (b) Initial subcategory averages for each seed. (c) Objective function  $J(\Theta)$ convergence per epoch. (d) Final subcategory averages after optimization; some skew remains due to inherent data imbalances, but all seeds converge to the same thresholds}
    \label{fig:cem_results}
\end{figure*}

Consequentially, we need to approximate the gradient $\hat{\nabla} J(\Theta)$ using a successive approximation-based method.
Such methods perform gradient descent on the parameters $\Theta$ in an iterative manner, beginning with an initial guess $\Theta_0$.
At each iteration $k$, the parameters are updated using $\Theta_{k+1} = \Theta_{k}+\eta \hat{\nabla} J(\Theta_{k})$ where $\eta$ is the learning rate, and $\hat{\nabla} J(\Theta_{k})$ is an approximation of the gradient of $J(\Theta_{k})$.
Iterating using this successive approximation yields the optimal values of the thresholds, $\Theta^*$.
TAP uses the cross-entropy method~\cite{CEMethod} for approximating the gradient $\hat{\nabla} J(\Theta_{k})$.
The cross-entropy method approximates the gradient $\hat{\nabla} J(\Theta_{k})$ using finite differences:
\begin{equation}
    \hat{\nabla} J(\Theta_k) = \frac{J(\Theta_k+\epsilon)-J(\Theta_k-\epsilon)}{2 \epsilon}
\end{equation}
where, $J(\Theta_k-\epsilon)$ and $J(\Theta_k+\epsilon)$ are the values of the objective function computed in the $\epsilon$ neighborhood of $\Theta$.
$\epsilon$ can be adjusted based on the spread of the distribution $D$.
Although cross-entropy methods can be used to find all values of $\Theta$ concurrently, our data has three disjoint data distributions, $D_a$, $D_v$, and $D_{\omega}$, and
therefore, each threshold $\Theta_a$, $\Theta_v$, and $\Theta_{\omega}$ can be computed from its own optimization.
For example, $\Theta^*_{\omega} = \argmin_{\Theta_{\omega}} J(\Theta_{\omega})$.
As thresholds $\Theta_a$, $\Theta_v$, and $\Theta_{\omega}$ have physical meaning in our case - acceleration, velocity, and yaw rate - the initial guesses $\Theta_0$ can be based on a combination of domain knowledge and the data distributions $D_a$, $D_v$, and $D_{\omega}$.
Figure~\ref{fig:cem_results}(a) shows the spread of these distributions along with initial threshold guesses.

\section{Experiments and Results}
We evaluate TAP by verifying the effectiveness of cross-entropy in learning optimal thresholds, comparing extracted SDL label accuracy with two baselines, and demonstrating TAP’s ability to identify unique trajectories in the data.

\subsection{Dataset and Preprocessing}
We use the publicly available Waymo Open Motion Dataset (WOMD)~\cite{WOMD_2021} as our source of scenarios \( S \). 
TAP is evaluated on a representative subset of 492 scenarios from WOMD, containing $25,889$ vehicle trajectories, each recorded at 10 Hz over a duration of \( T = 9.0 \) seconds.
Each scenario \( S_i \) includes data from several vehicles \( V^i_j \) including their states \(\{x, y, v, \phi\}\), where \(x, y\) denote the vehicle's position in map coordinates, \( v \) represents its velocity in m/s, and \( \phi \) is the heading in radians.
To obtain the complete state vector \( s = \{x, y, v, a, \phi, \omega\} \), we derive acceleration \( a \) and yaw rate \( \omega \) from the recorded velocity and heading. 
Each vehicle’s trajectory is normalized to a common initial condition by translating positions \( x, y \) to the origin and adjusting the heading to \( \phi = 0 \) radians, enabling fair trajectory comparisons.
Using the full state vector \( s = \{x, y, v, a, \phi, \omega\} \), we construct 1-second average distributions \( D_a \), \( D_v \), and \( D_{\omega} \) (Section~\ref{sec:problem_threshold}), which serve as the foundation for TAP’s automatic threshold computation.
WOMD trajectories do not contain ground truth action labels.

\subsection{Result 1 - Automated Threshold Optimization}
The first result is to evaluate the effectiveness of the TAP in automatically extracting SDL labels from trajectory data. 
Specifically, we evaluate if the cross-entropy optimization can effectively learn the values for thresholds for yaw rate (\(\Theta_\omega\)), acceleration (\(\Theta_a\)), and velocity (\(\Theta_v\)) from the corresponding data distributions.


Sensor noise in trajectory data complicates distinguishing between truly stopped vehicles and those moving at very low speeds.
Setting the stop threshold $\theta_{v,\text{stop}}$ too low risks misclassifying stationary vehicles, while a threshold set too high may incorrectly label slow-moving vehicles as stopped. 
To mitigate this, we manually define $\theta_{v,\text{stop}} = 0.1 m/s$, reducing the impact of sensor fluctuations while maintaining accurate classification. 
The remaining thresholds in $\Theta$ are optimized using the cross-entropy method.
Figure~\ref{fig:cem_results} presents the results of threshold optimization, with Figure~\ref{fig:cem_results}(a), showing the 1-second data distributions for $D_a$, $D_v$, and $D_{\omega}$.

\begin{figure*}
    \centering
    \includegraphics[width=\linewidth]{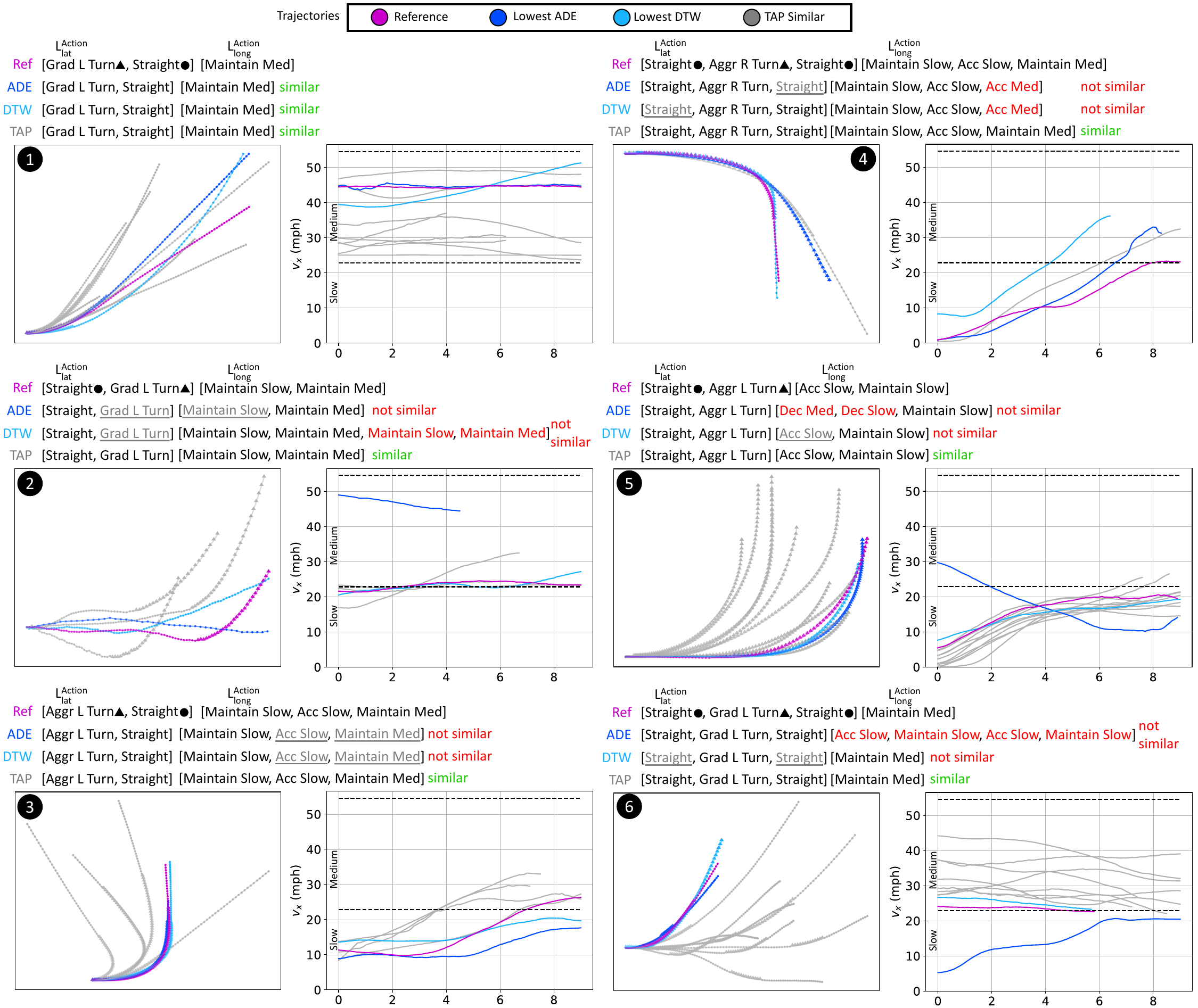}
    \caption{Trajectories with similar SDL labels to the reference (magenta). The trajectory with the lowest ADE is shown in dark blue, and lowest DTW in light blue. Red labels indicate extra actions, while underlined labels indicate missing actions. 
    Vehicle velocity plots are shown on the right.}
    \label{fig:simsearch_results}
\end{figure*}

\noindent \textbf{Hyperparameters and Ablation:}
The cross-entropy method requires initial threshold seeds $\Theta^1_0, \Theta^2_0, \Theta^3_0$ to ensure effective convergence.
We selected these seeds by analyzing the data distribution as well as using domain knowledge to achieve a starting point guess at balanced dataset partitions for cross-entropy.
To enhance robustness and avoid local minima, three distinct seeds were used for each threshold.
This approach improves data separability and ensures more reliable SDL extraction. 
The hyperparameters are listed in Table~\ref{table:hyperparameters}.
Figure~\ref{fig:cem_results}(a) also shows these initial threshold guesses plotted for their respective distributions.
These thresholds define initial intra-partition separation values (Figure~\ref{fig:cem_results}(b)), $\mu_{part}$, which measure the distance between samples within each partition with higher values indicating greater separation. 

\noindent \textbf{Evaluation:}
The cross-entropy method aims to optimize these thresholds for balanced separation across partitions.
Figure~\ref{fig:cem_results}(c) illustrates the convergence of the objective function \( J(\Theta) \) over multiple epochs, with similar final values for each seed indicating optimal thresholds. 
As \( J(\Theta) \) decreases, it demonstrates effective learning by the method. 
Figure~\ref{fig:cem_results}(d) shows the final intra-partition separations, where $\mu_{part}$ is more balanced across partitions compared to the initial values (Figure~\ref{fig:cem_results}(b)). 
The lateral thresholds converged to values of $\Theta_{\omega} = \{0.0283, 0.0754, 0.1541\}$ rad/s. The longitudinal thresholds converged to $\Theta_{a} = \{-1.3715, 1.5557\}$ m/s$^2$ and $\Theta_{v} = \{0.1, 10.2140, 24.4046\}$ m/s.
Consistent convergence across all seeds confirms the cross-entropy method’s reliability in optimizing partition balance.
\begin{table}
\centering
\renewcommand{\arraystretch}{1}
\begin{tabular}{ |c|c|c|c| }
 \hline
 \noalign{\vspace{0.1em}} 
$\downarrow$ \textbf{Hyperparameter} / Objective $\rightarrow$ & $\hat{J(\Theta_{a})}$ & $\hat{J(\Theta_{v})}$ & $\hat{J(\Theta_{\omega})}$ \\
 \hline
 \textbf{$\epsilon$} & 0.05 & 0.05 & 0.005 \\
\hline
 \textbf{$\eta$} & 0.05 & 0.2 & 0.01 \\
 \hline
\end{tabular}
\caption{The hyper-parameters $\epsilon$ and learning rate used for each category of the cross-entropy method.}
\label{table:hyperparameters}
\end{table}

\subsection{Result 2 - SDL Similarity Search}

We use TAP to extract SDL labels for each trajectory, as outlined in Section~\ref{sec:methodology_tap}. 
The objective of SDL similarity search is to find trajectories with similar behaviors by measuring the distance between their respective SDL labels.
We use the distance function $g_{\text{SDL}}(L_{\text{ref}},L^i_j) = 0$ if $L_{\text{ref}} = L^i_j$, and $g_{\text{SDL}}(L_{\text{ref}},L^i_j) = 1$ otherwise, which considers two trajectories similar if they share the same SDL label.
Other distance metrics, such as Levenshtein distance, could also be used for comparing SDL labels.
WOMD trajectories do not contain ground truth labels for similarity, therefore, we compare TAP's SDL-labels against two commonly used measures for trajectory similarity, Average Displacement Error (ADE)~\cite{ADE_similarity1,ADE_similarity2} and Dynamic Time Warping (DTW)~\cite{DTW_similarity1,DTW_similarity2}.
These measures are used to compare distance between trajectories in autonomous driving prediction, measuring similarity through the Euclidean distances between corresponding points along trajectories.

Figure~\ref{fig:simsearch_results} demonstrates TAP’s effectiveness in identifying behaviorally similar trajectories compared to baseline measures. 
Given a reference trajectory  $S^L_{\text{ref}}$ (magenta), TAP retrieves all scenarios with matching SDL labels (gray). 
The most similar trajectories identified by ADE (dark blue) and DTW (light blue) are also shown, along with TAP’s SDL labels for those trajectories.
In Figure~\ref{fig:simsearch_results} example \circled{1}, the lateral motion of the reference trajectory includes a gradual left turn (triangular marker) followed by straight (circular marker) and the longitudinal behavior is at medium speed. 
This is an example where trajectories returned by both ADE and DTW also match this behavior, yielding trajectories with the same SDL labels as TAP.
However, the other examples reveal cases where ADE and DTW fail to capture key behaviors. 
In \circled{2}, both baseline methods miss the turning behavior in the reference trajectory, while TAP identifies trajectories with similar behavior despite slight spatial differences. 
In \circled{3}, ADE and DTW match the lateral SDL label but fail to capture the longitudinal behavior in acceleration, highlighting TAP’s superior accuracy in comparing similar trajectories. 
With \circled{4}, the ADE trajectory starts straight and makes a similar aggressive right turn as the reference but fails to straighten afterward. In contrast, the nearest DTW trajectory begins its turn prematurely, missing the initial straight phase entirely. Additionally, both baseline trajectories fail to replicate the reference’s longitudinal behavior.
In \circled{5}, the nearest ADE and DTW trajectories again deviate significantly in longitudinal dynamics. The ADE trajectory begins at a medium speed and decelerates to a slow speed, while the DTW trajectory fails to accelerate enough to be classified as accelerating.
Example \circled{6} shows the two measures failing for different reasons. The ADE trajectory matches the lateral pattern but travels at a much slower speed. Conversely, the DTW trajectory mirrors the longitudinal behavior but transforms the entire sequence into a prolonged turn without straight-line segments.

For empirical comparison, we conducted similarity searches on 25,889 trajectories. 
ADE failed to retrieve behaviorally similar trajectories for 7,924 samples $(30.56\%)$, often selecting spatially close trajectories with differing behaviors. 
DTW exhibited similar issues, with 6,246 cases $(24.13\%)$ failing to match the reference’s behavior.
These examples demonstrate the limitations of using traditional measures for behavior-based similarity searches. 
ADE and DTW focus only on spatial proximity, which can miss critical driving behaviors. 
In contrast, TAP’s SDL-based approach excels in identifying behaviorally similar trajectories, enabling more reliable comparisons for robust AV testing and real-world scenario evaluation.
While this result focuses on comparing one trajectory at a time, our ongoing work extends this to multiple trajectory comparisons, enabling scenario-level analysis for more comprehensive behavior evaluations.

\noindent \textbf{Computation Time:} Although TAP is designed as an offline method, its execution time compares as follows: for 1,000 trajectory comparisons, TAP takes 3,831 ms, ADE 302 ms, and DTW 2,096 ms on a 2.1 GHz Intel Xeon (8 cores). TAP spends only 3 ms on comparisons, with 3,828 ms allocated to label generation.
Since the comparison time increases at $O(n^2)$ and the generation of labels at $O(n)$, the relative efficiency of TAP improves with larger datasets.
\label{sec:result2}

\subsection{Result 3 - Unique Behaviors}
\begin{figure}
    \centering
    \includegraphics[width=\linewidth]{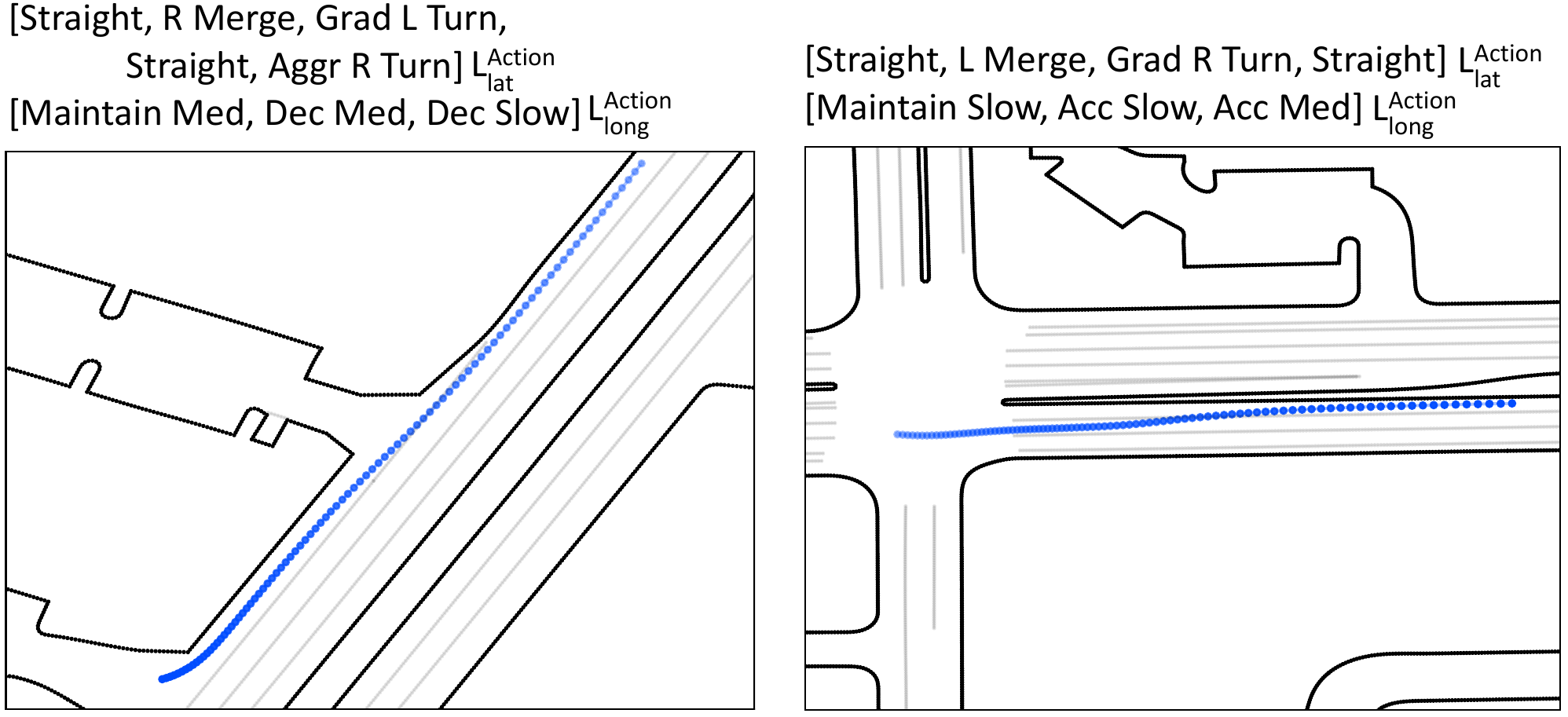}
    \caption{Two trajectories whose labels were completely unique within the dataset. Both have many lateral and longitudinal actions, which is uncommon among the dataset, resulting in unique samples.}
    \label{fig:unique_results}
\end{figure}

In addition to identifying similar behaviors, we used SDL label distances to detect unique trajectories, classified as such when no other trajectory in the dataset shares the same SDL label. 
From the Waymo Open Motion Dataset, we identified 395 unique trajectories out of 25,889, revealing rare or complex driving patterns.
Figure~\ref{fig:unique_results} shows two examples of unique trajectories.
The left depicts a vehicle entering a turn lane, making a right turn while decelerating. 
This trajectory is labeled unique due to five lateral and three longitudinal actions within a short 9-second window, suggesting a potentially complex or erratic driving pattern. 
The right example involves a vehicle changing lanes while accelerating, similarly marked as unique due to its dense sequence of actions. 
Although this result is for single trajectory similarity, these findings demonstrate that beyond trajectory similarity, detecting such unique trajectories can help uncover instances where an AV performed rare or unexpected maneuvers. 
This is crucial for identifying potentially risky or erratic behaviors, enabling better safety analysis and anomaly detection in large-scale AV datasets.



\section{Discussion and Conclusion}

This paper introduced the Trajectory-to-Action Pipeline (TAP), a method for automatically extracting interpretable Scenario Description Language (SDL) labels from trajectory data. TAP employs rule-based automata with data-derived thresholds, optimized using cross-entropy to generate structured, hierarchical labels. Using the Waymo Open Motion Dataset for validation, TAP demonstrated a 30\% improvement in precision over Average Displacement Error (ADE) and a 24\% improvement over Dynamic Time Warping (DTW) in identifying behaviorally similar trajectories.
TAP also enables the automatic detection of unique scenarios within large datasets.
Despite these contributions, there are key areas for future improvements. 
Currently, TAP focuses solely on vehicle trajectories, without integrating static environmental elements (e.g., road geometry) or accounting for interactions between multiple agents.
Future work would enhance TAP by incorporating these to support more complex scenario comparisons.
Additionally, while TAP introduces its own SDL schema, adapting the method to established frameworks like Scenic or OpenScenario could improve interoperability with existing testing and simulation tools.
Lastly, TAP could be augmented using recent transformer-based architectures for encoding trajectories.
Ultimately, TAP lays a strong foundation for automating behavior-based analysis in AV testing and safety evaluation, with potential for significant extensions to better reflect real-world scenario complexity.

\bibliographystyle{unsrt}
\bibliography{references,mb}


\end{document}